\documentclass[11pt]{article}

\usepackage[final]{acl}

\usepackage{times}
\usepackage{latexsym}

\usepackage[T1]{fontenc}

\usepackage[utf8]{inputenc}

\usepackage{microtype}

\usepackage{inconsolata}

\usepackage{graphicx}

\usepackage{booktabs}
\usepackage{multirow}
\usepackage{CJKutf8}
\usepackage{todonotes}
\usepackage{placeins}
\usepackage{float}
\usepackage{dblfloatfix}
\usepackage{tcolorbox}
\DeclareFontFamily{C19}{fallback}{}
\DeclareFontShape{C19}{fallback}{m}{n}{<-> simsun}{} 
\newcommand{\cobox}{\includegraphics[height=1.8ex]{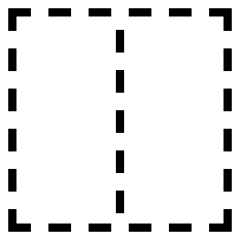}}
\newcommand{\zho}[1]{\begin{CJK*}{UTF8}{gbsn}#1\end{CJK*}}
\newcommand{\jpn}[1]{\begin{CJK*}{UTF8}{min}#1\end{CJK*}}
\newcommand{\kor}[1]{\begin{CJK*}{UTF8}{mj}#1\end{CJK*}}

\title{EXECUTE: A Multilingual Benchmark for LLM Token Understanding}

\author{Lukas Edman$^{1,3}$ \qquad Helmut Schmid$^{2}$ \qquad Alexander Fraser$^{1,3,4}$ \vspace{.2cm}\\ 
$^{1}$School of Computation, Information and Technology, TU Munich \\
$^{2}$Center for Information and Language Processing, LMU Munich \\ 
$^{3}$Munich Center for Machine Learning \\ 
$^{4}$Munich Data Science Institute \\ 
\vspace{.1cm} {\tt \small lukas.edman@tum.de, schmid@cis.lmu.de}
}

\begin{document}
\maketitle
\begin{abstract}
The CUTE benchmark showed that LLMs struggle with character understanding in English. We extend it to more languages with diverse scripts and writing systems, introducing EXECUTE. 
Our simplified framework allows easy expansion to any language.
Tests across multiple LLMs reveal that challenges in other languages are not always on the character level as in English. Some languages show word-level processing issues, some show no issues at all.
We also examine sub-character tasks in Chinese, Japanese, and Korean to assess LLMs' understanding of character components.

\end{abstract}

\section{Introduction}
LLMs perform well on many tasks but struggle when they are asked to manipulate character sequences, as shown by the CUTE benchmark \cite{edman-etal-2024-cute}. While CUTE tested Russian, showing this issue is not language-specific, it failed to consider other linguistic differences that may affect results.
Language variation extends beyond script differences to writing system differences. English and Russian use alphabets. Other languages use Abugidas, where letters are not strictly ordered within syllables, or Abjads, which mark vowels with diacritics or not at all. Chinese uses a logographic script, where most words are just 1-2 characters long. Multilingual LLMs allocate tokens unevenly across languages: high-resource languages are well represented, but some low-resource languages are mainly processed at the byte level.

\begin{table}[!htp]\centering
\scriptsize
\begin{tabular}{lrrrrrr}\toprule
Language &Script &Writing System &$c/w$ &$t/w$ &$c/t$ \\\midrule
Amharic &Ge'ez &Abugida &3.71 &7.69 &0.48 \\
Arabic &Arabic &Abjad &4.63 &2.43 &1.90 \\
Chinese &Simpl. Han &Logographic &1.51 &1.25 &1.20 \\
English &Latin &Alphabet &4.04 &1.32 &3.05 \\
Hindi &Devanagari &Abugida &3.66 &2.80 &1.31 \\
Japanese &Japanese &Mixed &1.54 &1.27 &1.22 \\
Korean &Hangul &Featural &3.38 &2.71 &1.25 \\
Russian &Cyrillic &Alphabet &5.06 &2.36 &2.14 \\
\bottomrule
\end{tabular}
\caption{CWT statistics of EXECUTE's languages. $c$, $w$, and $t$ denote characters, words, and tokens. $c/w$ refers to the average characters per word. $t$ is the average token count across the 5 tokenizers used by the models.
}\label{tab:langs}
\end{table}

\begin{figure}
    \centering
    \includegraphics[width=\linewidth]{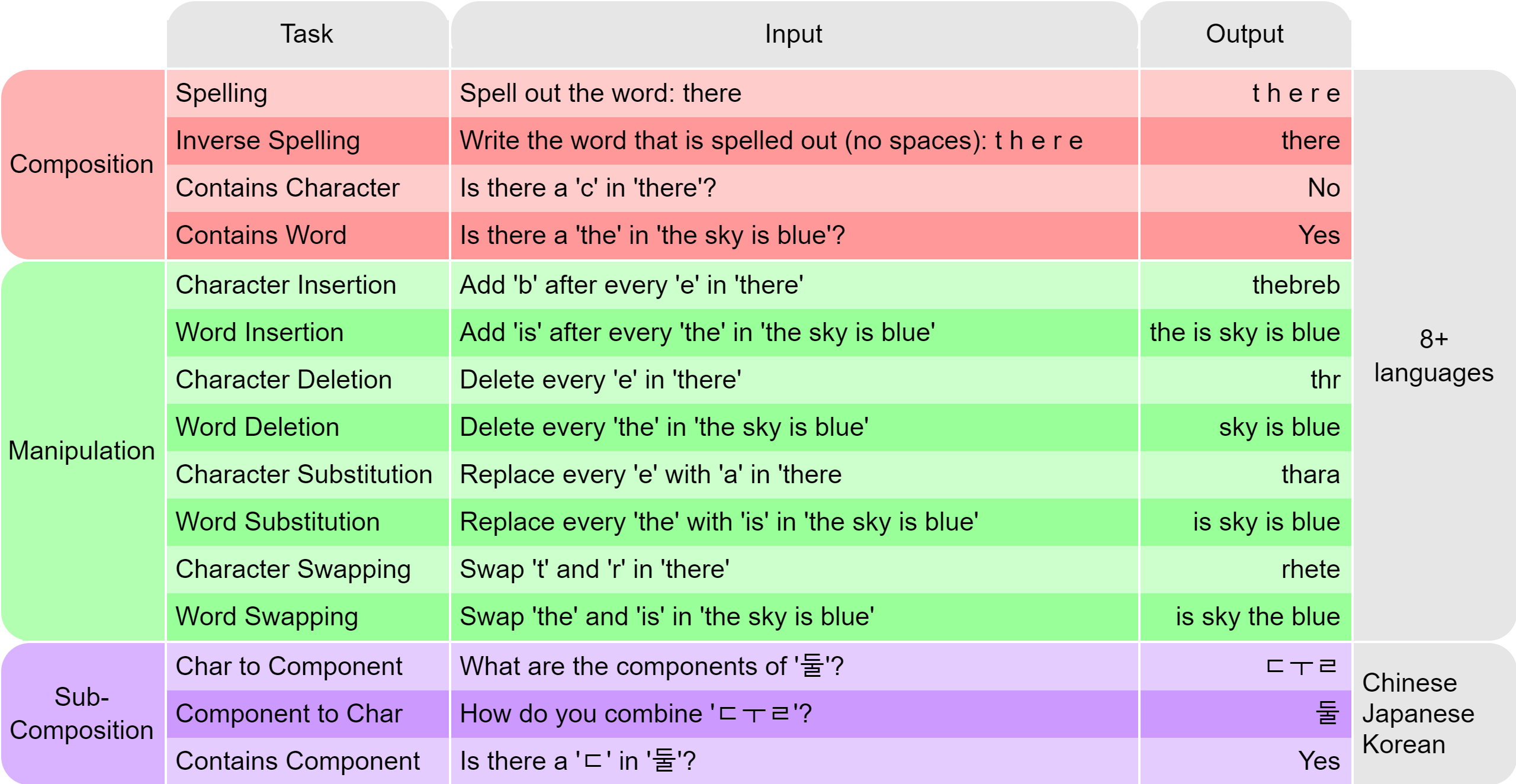}
    \caption{EXECUTE benchmark. Prompts shortened for brevity. Example of full prompt in Appendix \ref{app:prompt}.}
    \label{fig:tasks}
\end{figure}

We explore these languages in our benchmark EXECUTE: the \textbf{E}xpandable \textbf{X}(Cross)-Lingual \textbf{E}xtension of \textbf{CUTE}.\footnote{\url{https://github.com/Leukas/EXECUTE}} We mainly look at 8 languages, shown in Table \ref{tab:langs}, which vary in script, writing system, tokenization, and resourcedness. We also provide a framework for adding languages to make this benchmark easily expandable.  
In our results and analysis, we find that:
\begin{enumerate}
    \itemsep-3pt
    \item Benchmark results for non-English languages often differ from the English results.
    \item The results correlate with the languages' CWT (character-word-token) statistics (see Table \ref{tab:langs}).
    \item Surprisingly, the less an LLM knows a language, the better it performs on EXECUTE.
    \item LLMs struggle with understanding sub-character components (see Figure \ref{fig:tasks}).
\end{enumerate}
Our results provide more insight into how LLMs process tokens on different granularities.

\section{Related Works}

Our work builds upon the CUTE benchmark \cite{edman-etal-2024-cute}, which showed that LLMs struggle with character manipulation tasks. CUTE was mainly created for English but also included Russian tasks, showing similar results. Similar studies probe models to spell or modify text on the character level, but either first train the model \cite{itzhak-levy-2022-models,kaushal-mahowald-2022-tokens}, or focus on other topics than orthography \cite{huang-etal-2023-inducing,efrat-etal-2023-lmentry}.

Research on error correction, including spelling correction, has been done for many languages. \citet{maxutov-etal-2024-llms} found spelling correction to be ``hard'' for LLMs in Kazakh. \citet{li2023ineffectivenesslargelanguagemodels} reported that LLMs perform worse than fine-tuned models for Chinese spelling correction. Similarly, \citet{kwon-etal-2023-beyond} showed that fine-tuned models outperform prompted LLMs for Arabic.
Spelling correction requires both character-level and semantic knowledge to determine the correct replacement. EXECUTE, like CUTE, aims to remove contextual semantic understanding from the benchmark.

Our sub-character experiments build on work by \citet{wu2025impactvisualinformationchinese} who released a detailed analysis of the information in Chinese characters. Our character-to/from-radical tasks resemble theirs, but they focus on simplified Chinese, while we also examine traditional characters via Japanese Kanji.

Character-level LLMs have been proposed as a solution to CUTE and have been shown to outperform subword LLMs in \citet{pagnoni2024bytelatenttransformerpatches}.


\section{Benchmark}
Figure \ref{fig:tasks} exemplifies our EXECUTE benchmark. We use the same composition and manipulation tasks as CUTE but drop the similarity tasks which require static embeddings (such as word2vec) and fluent speakers to define similarity thresholds, which vary by language and lack clear criteria. 
Their removal makes EXECUTE easier to expand. 
Adding a new language $X$ now only requires an English$\rightarrow$$X$ translation system. As cross-language alignment is not crucial, translations do not need to be perfect: grammaticality is preferable but not necessary. 
We modify prompt examples and the dataset used, so English and Russian results differ from CUTE's.\footnote{As our changes are minor, users of the English and Russian datasets should cite \citet{edman-etal-2024-cute}.} 
Although perfect translations are not required, we have fluent speakers verify that most translations preserve meaning and grammar. Table~\ref{tab:langs} lists these languages, covering eight major scripts and all known writing systems. While some widely used languages (e.g. Spanish) are missing, their script is represented, 
and Appendix~\ref{app:lang_sim} shows the performance of languages using the same script is highly correlated. 

We keep the prompt texts in English but use language-specific examples, since 
fully Russian prompts did not improve performance for Russian \cite{edman-etal-2024-cute}. It also ensures 
that the LLMs understand the task consistently across languages. 


\subsection{Data Preparation}
We now describe the exact differences in preprocessing steps between our benchmark and the CUTE benchmark. Although there are several changes, we find that the scores from CUTE and EXECUTE are still largely comparable, as shown in Appendix \ref{app:cute_comp}.

To start, we use an updated subset of 5000 stories from the TinyStories dataset \cite{eldan2023tinystoriessmalllanguagemodels}, which used GPT-4 to produce outputs rather than the GPT-3.5 outputs, used in CUTE. We find this dataset to be cleaner (with no random foreign characters), and it is purported by the dataset authors to also be of higher quality. For non-English languages, we translate all the stories using Google Translate. At this point, for Chinese and Japanese, it is necessary to apply word segmentation. For Chinese, we use \texttt{jieba}\footnote{\url{https://github.com/fxsjy/jieba}}, and for Japanese we use \texttt{nagisa}.\footnote{\url{https://github.com/taishi-i/nagisa}}

We then generate a character set and vocabulary from the translated stories to use for our tasks. This is unlike what is used in CUTE, which 
predefined alphabets and vocabularies taken from the Trillion Word Corpus and Wikipedia. This change is necessary as it is more difficult to define a strict character set for some languages, and also more difficult to find a vocabulary. 

In the CUTE benchmark, the vocabulary also removed words less than 3 characters to maintain a level of difficulty for the tasks. We remove this cutoff for Chinese, Japanese, and Korean, as it is too restrictive. 

As the prompts are few-shot, we require language-specific examples in each prompt. For CUTE, these examples were created manually. Instead, we generate 4 additional examples in the same manner as our test set, with a few additional stipulations:
\begin{itemize}
    \itemsep-3pt
    \item At least 2 examples must use a word that contains duplicate letters.
    \item At least 1 example must operate on the duplicate letters when applicable.
    \item For the contains tasks, 2 examples must have the label ``yes'' and 2 ``no''.
\end{itemize}

We specify the duplicate letter restrictions so that the LLM understands that it must modify \textit{all} of the targeted characters. 
The first two restrictions were not applied for Chinese however, as it is exceedingly rare for a Chinese word to contain duplicate characters. The last restriction is intended to ensure the model is not biased to answering either ``yes'' or ``no'' due to its frequency in the examples, a phenomenon which has been shown to be problematic in \citet{zhao2021calibrate}.

\paragraph{Diacritics}
Abugidas such as Hindi have diacritics to mark vowel sounds, aspirations, and nasalizations. Due to the complex rules surrounding valid diacritics, which also vary between languages, we opt to consider each ``character'' as the letter plus any diacritics attached, also known as the grapheme. This is already the case for Amharic, as the diacritics have become fused with consonants in the Ge'ez script itself.

\begin{figure*}[htpb]
    \centering
    \includegraphics[width=\linewidth]{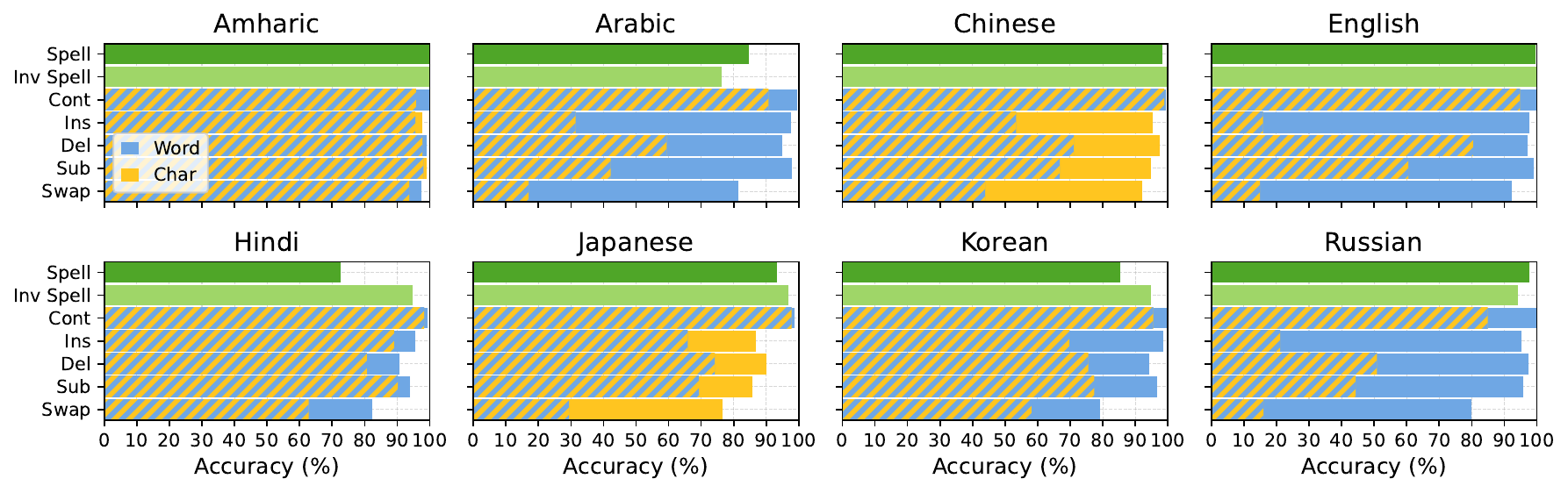}
    \caption{The best result of all models for each language and task.}
    \label{fig:best_main}
\end{figure*}

\subsection{Sub-Character Experiments}
Chinese, Japanese, and to a lesser extent Korean, have few characters per word, so we add language-specific tasks to assess their understanding of character components.

In Chinese, each character can be broken down into parts known as Kangxi radicals. An example of a decomposition is: \zho{晚} $\rightarrow$ \cobox \zho{日 免}, where {\cobox} indicates that \zho{日} should be placed to the left of \zho{免}. The radicals often have a related meaning to the composite: 
\zho{晚} means \textit{evening}, \zho{日} means \textit{sun} and \zho{免} means \textit{avoid}. Japanese Kanji characters originate from traditional Chinese characters and can also be decomposed into radicals.
Korean Hangul characters denote syllables and can be split into Jamo, which correspond to 
phonemes. 
For example, \kor{둘} (\textit{dul}) becomes \kor{ㄷ}(\textit{d}), \kor{ㅜ}(\textit{u}), and \kor{ㄹ}(\textit{l}).




We test the LLMs' ability to compose and decompose CJK characters into their components. For Chinese and Japanese, we ask the model to split characters into Kangxi radicals, and vice versa.\footnote{One can further split Kangxi radicals down to strokes, but this showed very poor performance in initial tests.} 
Similarly, we decompose Hangul characters to Jamo and vice versa. These tasks are analogous to the 
\texttt{spelling} and \texttt{inverse\_spelling} tasks. We further add a task (similar to \texttt{contains}) which asks if a character contains a Kangxi radical or Jamo.

Japanese can either be written with Kanji characters or with phonetic Hiragana characters. We test LLMs' ability to convert Kanji in Appendix \ref{app:furi}.

\subsection{Models}
We test 5 
popular open-source multilingual LLMs: Aya Expanse, Gemma 2, Llama 3.1 and 3.3, Qwen 2.5, and Mistral \cite{dang2024aya,gemmateam2024gemma2improvingopen,dubey2024llama,qwen2025qwen25technicalreport,jiang2023mistral}. Their sizes range from 7B to 70B parameters, and 
their vocabularies contain between 128k and 256k tokens. 

\section{Results}

\begin{table}[!htp]\centering
\scriptsize
\begin{tabular}{lrrrrrrr}\toprule
& & & &\multicolumn{3}{c}{Eng} \\\cmidrule(lr){5-7}
&Amh &Tzm &Sat &Cipher &Byte &Reg \\\midrule
Spell &96.3 &100.0 &97.6 &100.0 &85.0 &99.5 \\
Inv Spell &99.8 &100.0 &99.3 &100.0 &0.0 &99.6 \\
Cont Char &91.8 &100.0 &98.0 &98.6 &82.8 &75.7 \\
Cont Word &99.6 &99.2 &98.9 &99.0 &96.7 &99.9 \\
Ins Char &97.8 &97.8 &98.2 &98.6 &20.9 &13.5 \\
Ins Word &92.8 &94.3 &91.9 &97.1 &1.4 &96.6 \\
Del Char &97.6 &99.7 &98.7 &98.9 &78.8 &67.5 \\
Del Word &97.6 &76.2 &88.8 &95.6 &3.7 &96.5 \\
Sub Char &96.6 &98.4 &98.3 &95.5 &61.5 &51.4 \\
Sub Word &96.2 &96.6 &90.1 &98.4 &5.9 &98.5 \\
Swap Char &93.7 &97.6 &92.8 &98.3 &29.0 &12.7 \\
Swap Word &97.3 &87.9 &90.0 &95.9 &6.6 &90.9 \\ \midrule
Avg &96.4 &95.6 &95.2 &98.0 &39.4 &75.2 \\
\bottomrule
\end{tabular}
\caption{Llama 3.3 on low-resource languages.}\label{tab:lr_langs}
\end{table}

We first examine results by language, showing the best model performance for each in Figure \ref{fig:best_main}.\footnote{We show the results per task, as well as results for Aya and Mistral, in Appendix \ref{app:full_res}.} Russian and Arabic results resemble English results. 
Hindi and Korean perform better at the word level than the character level, though the gap is smaller than for English, with stronger results in character-level insertion and swapping.
Japanese and Chinese perform better on the character level, which is expected since each character is a word or almost a word. However, word-level tasks may simply be harder in these languages, as they require modifying multiple tokens instead of just one.

\subsection{Amharic and Low-Resourcedness}
Amharic stands out from the rest of the results in that the performance is nearly perfect in the best-case scenario. This is particularly surprising as Amharic is the lowest-resource language of the 8, and most characters are split into bytes by the tokenizers, meaning each character is represented by 3 tokens. We suspect that the good performance might actually be \textit{because} of this low-resourcedness. As seen in \citet{edman-etal-2024-cute}, and also observed in this work, LLMs are biased to generating real words and grammatical sentences. Their lack of understanding of Amharic 
might weaken this bias.

\begin{table}[!htp]\centering
\scriptsize
\begin{tabular}{lccccccc}\toprule
&\multicolumn{2}{c}{Gemma 2} &\multicolumn{2}{c}{Llama 3.1} &Llama 3.3 &\multicolumn{2}{c}{Qwen 2.5} \\\cmidrule(lr){2-3}\cmidrule(lr){4-5}\cmidrule(lr){6-6}\cmidrule(lr){7-8}
&9B &27B &8B &70B &70B &7B &32B \\\midrule
Amh &80.5 &85.3 &75.7 &\textbf{95.9} &96.4 &41.9 &74.4 \\
Ara &51.6 &62.3 &52.1 &68.1 &67.8 &47.2 &\textbf{68.6} \\
Zho &70.2 &74.4 &71.3 &81.1 &79.7 &70.4 &\textbf{83.6} \\
Eng &64.8 &71.6 &61.9 &75.7 &75.2 &62.1 &\textbf{77.3} \\
Hin &47.9 &47.1 &43.8 &54.0 &56.4 &43.5 &\textbf{86.2} \\
Jpn &60.1 &65.2 &58.8 &73.1 &74.6 &62.1 &\textbf{77.9} \\
Kor &73.6 &\textbf{80.8} &62.1 &76.9 &76.1 &60.2 &\textbf{80.8} \\
Rus &53.6 &62.6 &51.0 &67.8 &67.8 &52.1 &\textbf{71.2} \\ \midrule
Avg &62.8 &68.7 &59.6 &74.1 &74.3 &54.9 &\textbf{77.5} \\
\bottomrule
\end{tabular}
\caption{Average score per language. Best in bold.}\label{tab:models}
\end{table}

\begin{table*}[!htp]\centering
\scriptsize
\begin{tabular}{lccccccccccccc}\toprule
& &\multicolumn{2}{c}{Aya Expanse} &\multicolumn{2}{c}{Gemma 2} &\multicolumn{2}{c}{Llama 3.1} &Llama 3.3 &\multicolumn{2}{c}{Qwen 2.5} &\multicolumn{2}{c}{Mistral} \\\cmidrule(lr){3-4}\cmidrule(lr){5-6}\cmidrule(lr){7-8}\cmidrule(lr){9-9}\cmidrule(lr){10-11}\cmidrule(lr){12-13}
& &8B &32B &9B &27B &8B &70B &70B &7B &32B &8B &24B \\\midrule
\multirow{3}{*}{Zho} &Char to Rad &0.0 &2.0 &0.0 &0.8 &0.0 &1.4 &1.8 &1.4 &\textbf{16.4} &0.8 &3.5 \\
&Rad to Char &1.4 &8.2 &2.5 &0.8 &2.5 &10.7 &11.9 &7.6 &\textbf{22.8} &2.5 &8.2 \\
&Contains Rad &55.4 &65.5 &\textbf{81.1} &79.3 &69.4 &73.5 &72.9 &68.0 &78.8 &62.2 &74.5 \\ \midrule
\multirow{3}{*}{Jpn} &Char to Rad &0.0 &0.7 &0.7 &0.0 &0.0 &0.0 &2.2 &2.2 &\textbf{9.2} &0.4 &3.7 \\
&Rad to Char &2.6 &8.5 &2.2 &0.4 &1.9 &13.7 &13.3 &5.2 &\textbf{20.3} &1.5 &7.4 \\
&Contains Rad &57.9 &61.6 &\textbf{86.4} &72.7 &69.7 &73.1 &76.0 &73.4 &76.0 &65.7 &83.4 \\ \midrule
\multirow{3}{*}{Kor} &Hangul to Jamo &7.5 &48.1 &54.6 &65.3 &24.5 &48.8 &45.6 &24.7 &57.4 &35.6 &\textbf{66.4} \\
&Jamo to Hangul &24.7 &49.0 &47.2 &\textbf{63.9} &41.7 &24.5 &24.3 &25.9 &42.2 &28.1 &51.5 \\
&Contains Jamo &63.7 &76.2 &93.2 &\textbf{96.6} &92.1 &87.5 &90.3 &75.3 &93.4 &73.2 &88.7 \\
\bottomrule
\end{tabular}
\caption{Sub-character-level results on CJK languages. Best in bold.}\label{tab:extra}
\end{table*}

We provide further evidence that language knowledge inversely correlates with EXECUTE performance by adding two low-resource languages, Tamazight and Santali. Their unique scripts (Tifinagh and Ol Chiki) are not used by any higher resource languages, forcing LLM tokenizers to operate at the byte level. These languages were likely seen rarely, if ever, during training. We compare results on Amharic's best-performing model, Llama 3.3.
Additionally, we test two variations of English: one encodes text using a cipher that maps Latin to Amharic characters, and the other forces the inputs to be byte-level (retaining the Latin alphabet). These experiments assess whether byte-level operation alone improves performance or if eliminating English recognition via ciphering is also necessary. We expect ciphered English to perform similarly to Amharic.
Table \ref{tab:lr_langs} shows that Llama achieves near-perfect results in the low-resource languages, as well as the ciphered English. 
Byte-level English improves character tasks but fails on word tasks, partly due to bias. The model rarely sees English at the byte level except in social media, leading to random casing and antspeak (extra spacing) in the output. Degenerate output (e.g. ``1 1 1 ...'') also occurs. Some fine-tuning with this byte-level approach would likely increase performance considerably.

Does Amharic's near-perfect score mean character- and word-level processing is solved? No, it shows LLMs can perform arbitrary manipulations but are hampered by their language understanding. As training data increases, Amharic performance will likely decline. So, this benchmark should complement standard NLU benchmarks for a complete assessment.

\subsection{Language Clusters}
Table \ref{tab:langs} groups languages with similar CWT statistics into five categories: 1) Arabic \& Russian, 2) Hindi \& Korean, 3) Japanese \& Chinese, 4) Amharic, and 5) English. Their similar benchmark performance suggests that segmentation, whether natural or from tokenization, impacts results. As expected, the statistics of Tamazight (4.37, 8.83, 0.49) and Santali (3.54, 8.38, 0.42) closely align with Amharic.

\subsection{Model Performance}
Table \ref{tab:models} shows model performance. Larger models generally perform better. However, this trend does not hold across model families, as Qwen 2.5 (32B) outperforms the larger 70B Llama 3 models. Llama 3.3, despite its stronger performance than Llama 3.1 on standard benchmarks, performs similarly here.

\citet{edman-etal-2024-cute} found that more training data improved results on CUTE, but we find no such trend. Among 7-9B models, Gemma was trained on 8T tokens, Llama on 15T, and Qwen on 18T \cite{gemmateam2024gemma2improvingopen, qwen2025qwen25technicalreport, dubey2024llama}, yet their performance is inversely correlated. While this may be coincidental, results on Amharic, Tamazight, and Santali raise doubts about whether more training data improves performance on this benchmark.

\subsection{Sub-Character Performance}
Table \ref{tab:extra} shows sub-character results. For Japanese and Chinese, models struggle to translate characters to and from their radical components but perform better on the \texttt{Contains} task, as it only requires identifying one radical. While some characters, like \zho{晚} (\textit{evening}), have components that clearly contribute to meaning, others are more ambiguous. For example, \zho{木} (\textit{tree}) is likely easier for models to identify in \zho{樟} (\textit{camphor tree}) compared to \zho{章} (\textit{chapter}, \textit{seal}).


LLMs are notably better at converting between Hangul and Jamo, likely due to Hangul's simpler structure or its more frequent decomposition in training data. However, the conversion still falls short of the near-perfect scores seen in the main \texttt{Spelling} and \texttt{Inverse Spelling} tasks.

\section{Conclusion}
We present a multilingual, multi-script extension of the CUTE benchmark to test token understanding in a variety of languages. The benchmark is designed to be easily expanded to new languages, allowing the token understanding of LLMs to be tested in any language. 
Our findings show that manipulation on the character level is challenging in some non-English languages, but word-level manipulation is challenging for some languages too. Understanding the components of characters in Chinese, Japanese, and Korean is also lacking. The performance of a language can be somewhat predicted by its character-word-token ratios. Surprisingly, LLMs perform better on lower-resourced languages, due to their knowledge of high-resourced languages acting as a bias against the benchmark's tasks. While \citet{edman-etal-2024-cute} hypothesized that character-level models would be promising for solving the CUTE benchmark, EXECUTE demonstrates an additional need for debiasing models so they can temporarily forget what they know about a language.

\section{Limitations}
We limit ourselves to 8 languages for the majority of this work. While we argue that the languages in the same scripts as the ones tested will likely have similar results, pointing to the correlation in results between English, Spanish, German, and Xhosa in Appendix \ref{app:lang_sim}, we cannot know for sure without testing them all. Several other scripts are not covered which may have differing performances. 

We also do not test very large language models above 70B parameters due to compute constraints. The CUTE benchmark added scores for the 405B parameter Llama 3.1 and found it made improvements across the board, but was still lacking on character-level insertion and swapping. We would expect similar improvements for our English results, but it is unclear how it would perform for other languages.

\section{Acknowledgments}
The work was supported by the European Research Council (ERC) under the European Union's Horizon Europe research and innovation programme (grant agreement No. 101113091) and by the German Research Foundation (DFG; grant FR 2829/7-1).


\bibliography{anthology,custom}

\appendix

\section{Comparison to CUTE} \label{app:cute_comp}
In Table \ref{tab:comparison}, we run Llama 3.1 8B on CUTE and compare the results to English and Russian EXECUTE. The results are very similar, with \texttt{Insert Word} appearing slightly easier in CUTE. This confirms that our changes did not dramatically alter any results.

\begin{table}[!htp]\centering
\scriptsize
\begin{tabular}{lrrrrr}\toprule
&\multicolumn{2}{c}{EXECUTE} &\multicolumn{2}{c}{CUTE} \\\cmidrule(lr){2-3} \cmidrule(lr){4-5}
&Eng &Rus &Eng &Rus \\\midrule
Spell &98.7 &\textbf{72.1} &\textbf{99.8} &64.5 \\
Inv Spell &96.2 &37.9 &\textbf{98.4} &\textbf{74.1} \\
Cont Char &65.1 &57.1 &\textbf{67.1} &\textbf{68.4} \\
Cont Word &\textbf{97.3} &\textbf{97.8} &86.8 &97.4 \\
Ins Char &\textbf{4.4} &6.7 &4.2 &\textbf{7.6} \\
Ins Word &48.2 &48.5 &\textbf{62.0} &\textbf{59.2} \\
Del Char &56.1 &33.1 &\textbf{56.6} &\textbf{43.7} \\
Del Word &76.2 &\textbf{91.8} &\textbf{83.7} &82.3 \\
Sub Char &\textbf{39.3} &29.0 &34.4 &\textbf{33.8} \\
Sub Word &\textbf{94.1} &\textbf{87.0} &90.4 &76.7 \\
Swap Char &\textbf{6.6} &4.8 &6.1 &\textbf{5.2} \\
Swap Word &60.4 &\textbf{46.5} &\textbf{63.7} &33.3 \\ \midrule
Average &61.9 &51.0 &\textbf{62.8} &\textbf{53.9} \\
\bottomrule
\end{tabular}
\caption{EXECUTE versus CUTE with Llama 3.1 8B.}\label{tab:comparison}
\end{table}

\section{Language Similarity} \label{app:lang_sim}
We conduct similarity tests to see how similar the trends are across languages. We conduct a Pearson correlation between two languages for each task for a given model and average the models' correlations together. We show the similarity of the languages as determined by the average correlation of the results from the 5 LLMs of size 7-9B in Table \ref{tab:main_sims}. The languages are not particularly similar to one another, apart from Japanese and Chinese (which share some characters) and Arabic and Russian. The similarity between Arabic and Russian is not entirely clear, though it could be that their ratios of characters-per-word and characters-per-token are quite similar (such is also the case for Japanese and Chinese).

\begin{table}[!htp]\centering
\scriptsize
\begin{tabular}{lrrrrrrrr}\toprule
&Ara &Zho &Eng &Hin &Jpn &Kor &Rus \\\midrule
Amh &0.64 &0.01 &0.66 &0.33 &0.23 &0.62 &0.60 \\
Ara & &-0.11 &0.76 &0.44 &0.13 &0.85 &0.92 \\
Zho & & &0.17 &0.65 &0.93 &0.26 &-0.06 \\
Eng & & & &0.45 &0.36 &0.77 &0.86 \\
Hin & & & & &0.75 &0.71 &0.38 \\
Jpn & & & & & &0.49 &0.17 \\
Kor & & & & & & &0.84 \\
\bottomrule
\end{tabular}
\caption{Average correlations between the results for each language pair.}\label{tab:main_sims}
\end{table}

We also correlate the results from English to other Latin-scripted languages, German, Spanish, and Xhosa, in Table \ref{tab:latin_sims}. Here we see the average correlation is at least 95\% between English, German, and Spanish, and at least 85\% to Xhosa. This suggests
that the results for other Latin-scripted languages will likely not deviate too much, even if the languages are distant in relation and differing in resourcedness.

\begin{table}[!htp]\centering
\scriptsize
\begin{tabular}{lrrrr}\toprule
&Deu &Spa &Xho \\\midrule
Eng &0.96 &0.95 &0.85 \\
Deu & &0.99 &0.90 \\
Spa & & &0.90 \\
\bottomrule
\end{tabular}
\caption{Average correlations between the results for Latin-scripted languages.}\label{tab:latin_sims}
\end{table}

\section{Japanese Furigana} \label{app:furi}
Aside from Kanji, Japanese has two other writing forms: Hiragana and Katakana. Typical Japanese text will use all three forms, with several words being a combination of Kanji and Hiragana, and even in rare cases, all three. While Kanji is logographic like Chinese, Hiragana and Katakana are syllabaries. Kanji and Hiragana are the most used, while Katakana is typically only used for foreign words or onomatopoeiae. As such, we focus on Kanji and Hiragana. All Kanji characters can be written as Hiragana, and Kanji is sometimes annotated with its corresponding Hiragana as a method of learning the pronunciation of Kanji characters. This practice is known as Furigana. So we use this Furigana method as a test in our benchmark, prompting the model to translate Kanji to Hiragana.\footnote{We do not do the reverse as multiple Kanji can have the same phoneme, e.g. \zho{考} and \zho{好} both denote \textit{ko}.} With this, we are essentially testing if the LLMs have a phonetic understanding of the Kanji.

\begin{figure}
    \centering
    \includegraphics[width=\linewidth]{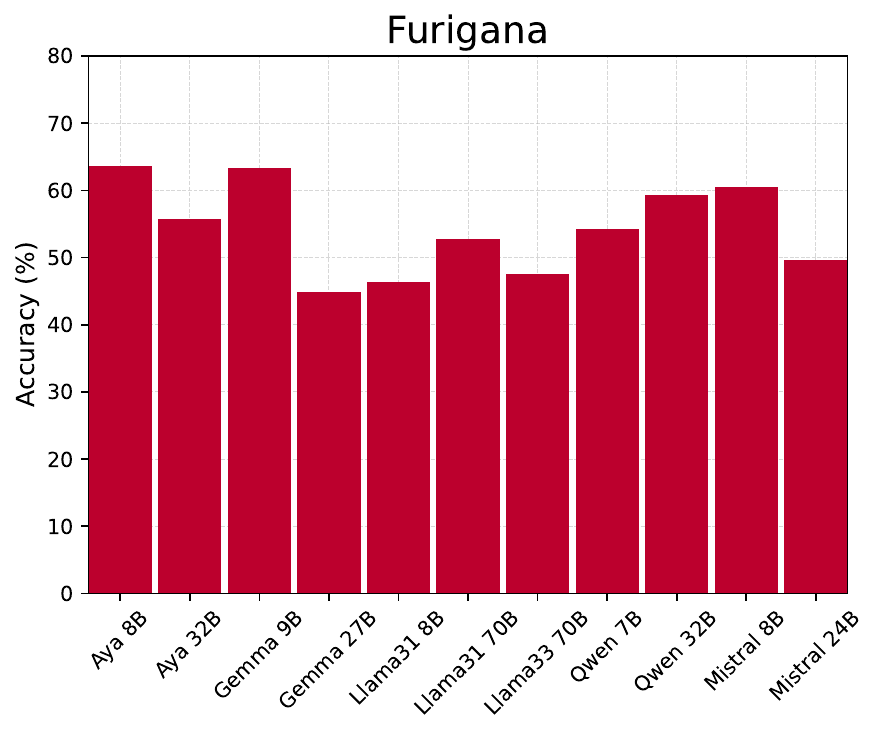}
    \caption{Performance on Kanji to Hiragana conversion.}
    \label{fig:furi}
\end{figure}


In Figure \ref{fig:furi}, we see the models' results on the Furigana task. 
Similar to the Korean Hangul to Jamo, the Kanji to Hiragana tasks show that the LLMs generally understand the task, but have not perfected it. Unlike the other sub-character tasks, converting from Kanji to Hiragana cannot be done purely visually. This requires knowledge of how a Kanji sounds, and which Hiragana denote which sounds. From this, we can see a partial understanding. 

\begin{figure}[!ht]
    \centering
    \small
\begin{tcolorbox}
\texttt{[INST] Spell out the word, putting spaces between each letter, based on the following examples: \\
\\ 
1. Spell out the word `` \jpn{かわいい} ''. Answer: `` \jpn{か わ い い} '' \\
2. Spell out the word `` \jpn{出し} ''. Answer: `` \jpn{出 し} '' \\
3. Spell out the word `` \jpn{応援} ''. Answer: `` \jpn{応 援} '' \\
4. Spell out the word `` \jpn{親友} ''. Answer: `` \jpn{親 友} '' \\
\\\raggedright
Question: Spell out the word `` \jpn{実行する} ''. [/INST] Answer: `` \jpn{実 行 す る} '' }
\end{tcolorbox}
    \caption{Example of full prompt for Japanese \texttt{spelling}, with intended output in red. \texttt{[INST]} and \texttt{[/INST]} denote any tokens added to enable normal behavior from each LLM. }
    \label{fig:prompt_ex}
\end{figure}

\section{Full Prompt Example} \label{app:prompt}
We show an example of a full prompt in Figure \ref{fig:prompt_ex}.

\section{Full Results} \label{app:full_res}
The complete results on EXECUTE for all the models tested are shown in Tables \ref{tab:full_res1} and \ref{tab:full_res2}.

\begin{table*}[!htp]\centering
\scriptsize
\begin{tabular}{llccccccccccc}\toprule
& &\multicolumn{2}{c}{Aya Expanse} &\multicolumn{2}{c}{Gemma 2} &\multicolumn{2}{c}{Llama 3.1} &Llama 3.3 &\multicolumn{2}{c}{Qwen 2.5} &\multicolumn{2}{c}{Mistral} \\\cmidrule(lr){3-4}\cmidrule(lr){5-6}\cmidrule(lr){7-8}\cmidrule(lr){9-9}\cmidrule(lr){10-11}\cmidrule(lr){12-13}
& &8B &32B &9B &27B &8B &70B &70B &7B &32B &8B &24B \\\midrule
\multirow{12}{*}{Amharic} &Spell &25.6 &72.4 &99.5 &91.2 &98.6 &98.4 &96.3 &0.9 &14.3 &97.0 &99.8 \\
&Inv Spell &77.6 &71.5 &99.1 &91.4 &98.8 &99.8 &99.8 &8.2 &52.4 &99.1 &100.0 \\
&Cont Char &58.5 &85.8 &73.0 &81.9 &90.4 &91.8 &91.8 &63.2 &93.5 &94.7 &95.8 \\
&Cont Word &55.9 &69.5 &97.0 &98.3 &78.5 &99.5 &99.6 &71.7 &99.9 &95.9 &98.8 \\
&Ins Char &35.2 &58.2 &57.1 &65.5 &26.1 &92.3 &97.8 &10.3 &57.9 &60.6 &92.2 \\
&Ins Word &70.7 &66.7 &95.6 &91.7 &28.7 &93.3 &92.8 &58.4 &92.6 &70.5 &87.9 \\
&Del Char &54.5 &85.0 &84.5 &89.4 &89.6 &96.8 &97.6 &43.5 &78.6 &85.3 &97.7 \\
&Del Word &85.8 &91.2 &63.1 &79.2 &93.2 &99.1 &97.6 &70.0 &91.1 &91.9 &89.4 \\
&Sub Char &63.6 &82.9 &70.1 &85.2 &87.5 &94.9 &96.6 &29.6 &67.3 &76.6 &99.1 \\
&Sub Word &91.9 &98.1 &98.1 &93.2 &91.6 &96.7 &96.2 &79.2 &90.9 &90.3 &96.5 \\
&Swap Char &20.2 &53.8 &53.0 &70.7 &52.8 &91.3 &93.7 &12.8 &65.8 &48.8 &86.4 \\
&Swap Word &60.7 &84.5 &75.7 &86.0 &73.0 &96.5 &97.3 &55.0 &88.1 &66.0 &94.9 \\ \midrule
\multirow{12}{*}{Arabic} &Spell &48.7 &74.1 &36.0 &69.9 &50.8 &84.7 &81.0 &20.7 &52.8 &19.3 &40.0 \\
&Inv Spell &48.4 &64.7 &48.3 &63.6 &44.9 &69.7 &63.0 &39.2 &76.4 &27.6 &60.9 \\
&Cont Char &63.9 &74.5 &70.3 &70.0 &70.6 &77.4 &76.4 &74.0 &90.7 &72.0 &78.6 \\
&Cont Word &88.1 &97.9 &99.0 &98.7 &96.9 &99.1 &99.1 &95.5 &99.4 &88.7 &99.4 \\
&Ins Char &13.7 &8.3 &7.6 &16.1 &2.9 &15.7 &17.8 &11.7 &31.4 &4.7 &12.5 \\
&Ins Word &35.8 &61.1 &90.8 &97.5 &51.2 &89.1 &96.4 &61.3 &96.3 &45.1 &86.4 \\
&Del Char &36.0 &56.4 &36.3 &45.5 &45.0 &55.8 &59.4 &40.7 &53.1 &20.8 &29.9 \\
&Del Word &74.0 &90.4 &64.6 &83.4 &92.5 &95.0 &88.6 &82.0 &91.6 &63.7 &88.8 \\
&Sub Char &17.6 &26.6 &20.7 &33.9 &24.3 &38.7 &36.0 &23.0 &42.2 &11.7 &20.7 \\
&Sub Word &72.7 &92.6 &95.0 &97.3 &90.8 &97.2 &98.0 &79.3 &95.4 &77.1 &92.2 \\
&Swap Char &5.9 &9.1 &4.5 &8.0 &8.1 &17.0 &16.7 &4.0 &14.3 &2.5 &7.5 \\
&Swap Word &26.3 &55.1 &46.3 &63.1 &47.6 &78.0 &81.3 &34.8 &79.8 &29.0 &62.0 \\\midrule
\multirow{12}{*}{Chinese} &Spell &83.2 &93.0 &84.3 &91.3 &93.6 &98.4 &98.0 &96.3 &98.2 &93.4 &90.9 \\
&Inv Spell &98.5 &97.3 &95.1 &96.2 &98.6 &98.4 &98.7 &98.9 &99.8 &98.8 &99.6 \\
&Cont Char &84.0 &97.2 &96.4 &95.4 &92.5 &97.3 &91.1 &98.7 &98.9 &96.0 &98.7 \\
&Cont Word &84.1 &99.0 &99.4 &98.8 &91.7 &95.1 &86.7 &94.3 &98.1 &94.2 &94.8 \\
&Ins Char &70.0 &57.2 &78.6 &81.4 &67.3 &90.6 &92.6 &73.6 &95.4 &53.2 &89.3 \\
&Ins Word &28.4 &33.2 &41.9 &53.0 &20.5 &46.5 &47.5 &43.7 &53.4 &34.1 &50.5 \\
&Del Char &79.6 &90.0 &85.4 &88.1 &86.8 &97.0 &97.6 &89.2 &97.3 &88.3 &94.8 \\
&Del Word &38.4 &57.3 &46.8 &56.9 &59.4 &71.1 &70.1 &54.0 &65.1 &53.2 &67.4 \\
&Sub Char &60.6 &68.4 &69.9 &75.1 &84.0 &94.2 &94.8 &80.0 &94.6 &75.0 &91.3 \\
&Sub Word &40.2 &54.9 &55.3 &64.7 &47.5 &56.3 &53.4 &43.1 &66.8 &46.1 &66.5 \\
&Swap Char &63.9 &71.9 &73.3 &69.8 &90.6 &92.0 &92.1 &62.6 &92.0 &75.4 &90.5 \\
&Swap Word &14.2 &26.6 &15.6 &21.9 &22.8 &36.1 &33.3 &10.7 &43.9 &15.1 &35.0 \\\midrule
\multirow{12}{*}{English} &Spell &96.7 &98.5 &99.3 &99.5 &98.7 &99.5 &99.5 &94.7 &98.6 &97.3 &99.0 \\
&Inv Spell &95.4 &98.5 &99.3 &99.6 &96.2 &99.8 &99.6 &98.3 &99.2 &91.0 &98.7 \\
&Cont Char &62.6 &73.1 &68.0 &69.5 &65.1 &80.3 &75.7 &81.9 &94.8 &66.7 &83.5 \\
&Cont Word &94.3 &99.4 &99.7 &99.7 &97.3 &100.0 &99.9 &98.1 &100.0 &93.3 &99.9 \\
&Ins Char &11.8 &6.0 &9.2 &7.8 &4.4 &10.9 &13.5 &7.1 &15.9 &7.4 &4.4 \\
&Ins Word &39.9 &60.6 &86.7 &96.8 &48.2 &94.9 &96.6 &70.2 &97.6 &51.9 &72.9 \\
&Del Char &35.0 &56.3 &58.5 &80.4 &56.1 &68.3 &67.5 &56.8 &70.5 &33.6 &72.4 \\
&Del Word &60.3 &77.5 &53.7 &77.7 &76.2 &97.1 &96.5 &78.5 &95.7 &74.3 &69.8 \\
&Sub Char &27.7 &42.2 &35.4 &60.5 &39.3 &53.5 &51.4 &29.0 &52.1 &27.7 &51.7 \\
&Sub Word &82.4 &96.1 &94.9 &97.9 &94.1 &98.1 &98.5 &92.9 &99.0 &92.9 &97.0 \\
&Swap Char &6.0 &9.8 &6.9 &8.9 &6.6 &14.9 &12.7 &3.2 &11.5 &6.8 &11.0 \\
&Swap Word &22.6 &64.5 &66.2 &60.8 &60.4 &91.0 &90.9 &34.7 &92.4 &42.2 &85.9 \\
\bottomrule
\end{tabular}
\caption{Results for Amharic, Arabic, Chinese, and English.}\label{tab:full_res1}
\end{table*}

\begin{table*}[!htp]\centering
\scriptsize
\begin{tabular}{llccccccccccc}\toprule
& &\multicolumn{2}{c}{Aya Expanse} &\multicolumn{2}{c}{Gemma 2} &\multicolumn{2}{c}{Llama 3.1} &Llama 3.3 &\multicolumn{2}{c}{Qwen 2.5} &\multicolumn{2}{c}{Mistral} \\\cmidrule(lr){3-4}\cmidrule(lr){5-6}\cmidrule(lr){7-8}\cmidrule(lr){9-9}\cmidrule(lr){10-11}\cmidrule(lr){12-13}
& &8B &32B &9B &27B &8B &70B &70B &7B &32B &8B &24B \\\midrule
\multirow{12}{*}{Hindi} &Spell &48.4 &69.4 &12.6 &16.0 &46.2 &72.5 &71.3 &20.5 &57.7 &23.4 &58.7 \\
&Inv Spell &76.8 &83.2 &71.9 &83.4 &76.0 &92.7 &92.9 &71.2 &94.6 &61.9 &87.2 \\
&Cont Char &68.4 &85.6 &75.1 &75.7 &72.6 &87.3 &83.2 &90.9 &98.3 &82.4 &88.9 \\
&Cont Word &89.4 &98.7 &95.7 &98.5 &93.3 &98.9 &93.0 &94.7 &99.3 &81.1 &94.8 \\
&Ins Char &41.0 &10.8 &25.8 &29.4 &15.8 &29.0 &32.9 &45.9 &89.0 &27.5 &36.5 \\
&Ins Word &6.3 &11.7 &77.3 &41.8 &13.6 &25.5 &47.6 &31.8 &95.5 &9.1 &16.4 \\
&Del Char &58.8 &66.5 &50.9 &67.6 &66.5 &76.9 &76.9 &50.9 &80.7 &44.3 &79.5 \\
&Del Word &6.6 &26.6 &26.6 &24.4 &40.6 &25.7 &31.7 &29.4 &90.8 &13.9 &19.7 \\
&Sub Char &45.7 &66.4 &38.2 &56.0 &42.1 &61.4 &63.6 &50.8 &90.2 &33.1 &65.7 \\
&Sub Word &3.7 &14.9 &62.3 &15.9 &19.1 &23.8 &25.4 &15.3 &93.8 &4.9 &17.0 \\
&Swap Char &14.5 &24.5 &8.4 &29.4 &19.5 &15.1 &22.2 &9.9 &62.8 &21.6 &33.6 \\
&Swap Word &2.1 &16.9 &30.6 &27.5 &20.1 &39.4 &35.7 &10.3 &82.2 &3.0 &11.5 \\\midrule
\multirow{12}{*}{Japanese} &Spell &52.9 &83.2 &68.5 &71.0 &73.2 &93.3 &92.2 &72.3 &86.5 &77.2 &84.5 \\
&Inv Spell &92.4 &88.4 &87.8 &90.5 &78.1 &96.9 &95.1 &90.4 &95.9 &88.5 &96.7 \\
&Cont Char &67.9 &91.6 &90.9 &88.8 &87.0 &95.2 &93.4 &93.7 &97.8 &86.7 &93.0 \\
&Cont Word &82.4 &92.7 &98.8 &97.3 &90.0 &89.4 &85.5 &89.9 &94.8 &84.6 &93.9 \\
&Ins Char &48.3 &31.5 &58.3 &68.8 &18.6 &72.1 &77.6 &69.7 &86.8 &29.3 &78.5 \\
&Ins Word &21.4 &34.2 &41.7 &61.0 &10.6 &42.1 &47.2 &44.0 &65.9 &30.4 &46.2 \\
&Del Char &62.0 &78.3 &60.5 &64.4 &79.8 &88.2 &90.0 &70.0 &84.3 &67.1 &86.5 \\
&Del Word &31.9 &60.4 &49.3 &52.5 &62.4 &68.5 &74.2 &52.3 &61.4 &38.0 &63.5 \\
&Sub Char &57.2 &68.8 &58.5 &66.3 &75.8 &83.2 &83.7 &64.4 &85.7 &59.3 &81.7 \\
&Sub Word &32.4 &55.1 &48.9 &58.8 &50.9 &62.8 &65.1 &50.9 &69.3 &38.5 &60.6 \\
&Swap Char &40.8 &52.0 &50.2 &49.2 &66.6 &63.9 &66.7 &40.4 &76.6 &45.7 &70.1 \\
&Swap Word &5.9 &13.2 &8.1 &13.3 &12.8 &21.9 &25.0 &7.5 &29.4 &9.3 &20.9 \\\midrule
\multirow{12}{*}{Korean} &Spell &43.6 &71.5 &67.5 &85.4 &51.5 &78.0 &73.4 &42.5 &66.6 &42.3 &56.1 \\
&Inv Spell &85.0 &93.8 &82.6 &86.8 &71.7 &88.9 &88.8 &84.2 &94.8 &64.6 &92.2 \\
&Cont Char &71.1 &84.5 &81.1 &86.2 &81.6 &85.3 &74.4 &90.7 &95.5 &76.4 &89.0 \\
&Cont Word &92.6 &97.2 &98.8 &99.0 &95.6 &99.4 &99.1 &96.7 &99.6 &84.9 &98.7 \\
&Ins Char &33.6 &20.6 &47.4 &54.6 &21.0 &52.5 &60.8 &39.6 &69.7 &20.2 &41.7 \\
&Ins Word &36.4 &72.0 &93.3 &98.6 &50.8 &91.8 &94.1 &66.8 &92.5 &45.0 &87.7 \\
&Del Char &44.7 &64.6 &69.1 &75.6 &62.5 &62.2 &63.2 &44.6 &64.1 &43.2 &56.6 \\
&Del Word &56.5 &81.8 &77.5 &88.6 &91.9 &94.3 &86.3 &76.7 &91.0 &75.4 &90.7 \\
&Sub Char &39.1 &62.0 &71.4 &77.4 &56.0 &65.5 &63.9 &50.4 &67.6 &37.9 &53.0 \\
&Sub Word &48.9 &78.7 &90.1 &96.7 &73.0 &91.2 &92.1 &68.8 &90.5 &65.1 &89.6 \\
&Swap Char &27.3 &33.3 &48.9 &47.5 &37.3 &42.0 &44.0 &31.4 &58.1 &20.0 &33.0 \\
&Swap Word &22.6 &48.3 &55.8 &73.6 &52.5 &71.1 &72.8 &29.9 &79.3 &27.1 &61.1 \\\midrule
\multirow{12}{*}{Russian} &Spell &40.9 &80.4 &54.5 &88.3 &72.1 &97.6 &96.7 &54.0 &88.9 &79.8 &94.8 \\
&Inv Spell &54.1 &78.7 &71.4 &90.3 &37.9 &86.6 &86.0 &81.9 &94.2 &39.5 &88.0 \\
&Cont Char &55.0 &68.8 &59.7 &58.7 &57.1 &60.2 &59.1 &75.3 &84.9 &63.3 &77.6 \\
&Cont Word &96.9 &98.1 &99.6 &99.8 &97.8 &99.8 &99.9 &95.8 &99.5 &94.7 &99.5 \\
&Ins Char &10.9 &3.8 &7.0 &9.4 &6.7 &11.0 &12.0 &7.2 &21.1 &5.7 &11.3 \\
&Ins Word &29.3 &61.9 &90.6 &95.4 &48.5 &90.7 &93.1 &77.8 &95.1 &62.9 &84.2 \\
&Del Char &12.7 &34.7 &20.9 &38.1 &33.1 &49.9 &50.9 &24.8 &44.7 &18.9 &45.1 \\
&Del Word &68.6 &85.0 &75.0 &81.0 &91.8 &97.3 &96.7 &83.1 &90.7 &80.1 &85.7 \\
&Sub Char &15.5 &26.0 &17.0 &35.3 &29.0 &38.7 &40.2 &22.6 &44.3 &16.2 &39.7 \\
&Sub Word &63.8 &86.2 &94.5 &95.1 &87.0 &95.6 &95.8 &80.8 &95.2 &86.5 &95.6 \\
&Swap Char &1.3 &4.8 &2.6 &4.0 &4.8 &12.5 &13.6 &1.4 &16.0 &4.4 &12.0 \\
&Swap Word &26.1 &48.1 &50.7 &55.8 &46.5 &74.3 &70.0 &20.7 &79.9 &32.6 &79.5 \\
\bottomrule
\end{tabular}
\caption{Results for Hindi, Korean, Japanese, and Russian.}\label{tab:full_res2}
\end{table*}

\end{document}